%% file: main.tex
\documentclass[11pt, a4paper, logo, copyright]{deepmind}

\usepackage[all]{xy}

\usepackage[sort&compress]{natbib}
\usepackage{import}
\graphicspath{{figures/}}

\usepackage{epigraph}

\usepackage{enumitem}
\setlist[itemize]{topsep=0pt}

\usepackage{xcolor}

\newcommand{\indep}{~\bot\!\!\!\bot~}
\newcommand{\DO}{\mathrm{do}}

\newtheorem{zrem}{Remark}
\newenvironment{rem}{\par\small\zrem}{\endzrem}

\title{Shaking the foundations: delusions in sequence models for interaction and control}

\correspondingauthor{pedroortega@deepmind.com}

\keywords{sequence models, sequential prediction, reinforcement learning, causality, self-delusion}
\paperurl{Technical Report}

\reportnumber{001} % Leave blank if n/a

\author[*]{Pedro A. Ortega}
\author[*]{Markus Kunesch}
\author[*]{Gr\'egoire Del\'etang}
\author[*]{Tim Genewein}
\author[*]{Jordi Grau-Moya}

\author[1]{Joel Veness}
\author[1]{Jonas Buchli}
\author[1]{Jonas Degrave}
\author[1]{Bilal Piot}
\author[1]{Julien Perolat}
\author[1]{Tom Everitt}
\author[1]{Corentin Tallec}
\author[1]{Emilio Parisotto}
\author[1]{Tom Erez}
\author[1]{Yutian Chen}
\author[1]{Scott Reed}
\author[1]{Marcus Hutter}
\author[1]{Nando de Freitas}
\author[1]{Shane Legg}

\affil[*]{Deepmind Safety Analysis}
\affil[1]{DeepMind}
\begin{abstract}
The recent phenomenal success of language models has reinvigorated machine learning research, and large sequence models such as transformers are being applied to a variety of domains. One important problem class that has remained relatively elusive however is purposeful adaptive behavior. Currently there is a common perception that sequence models ``lack the understanding of the cause and effect of their actions'' leading them to draw incorrect inferences due to auto-suggestive delusions. In this report we explain where this mismatch originates, and show that it can be resolved by treating actions as causal interventions. Finally, we show that in supervised learning, one can teach a system to condition or intervene on data by training with factual and counterfactual error signals respectively. 
\end{abstract}

\begin{document}

\maketitle

\setlength{\epigraphwidth}{0.5\textwidth}
\epigraph{%
\begin{itemize}
    \item[User:] What's on the menu?
    \item[AI:] Cheeseburgers are on the menu.
    \item[User:] No, it's tacos.
    \item[AI:] Tacos are on the menu.
    \item[User:] What's on the menu?
    \item[AI:] Cheeseburgers are on the menu.
    \item[User:] No, I said it was tacos.
    \item[AI:] Tacos are on the menu.
    \item[User:] What's on the menu?
    \item[AI:] Cheeseburgers are on the menu.
\end{itemize}
\leavevmode}%
{\emph{--- Imagined conversation with a fast food AI.}}%

%%%%%%%%%%%%%%%%%%%%%%%%%%%%%%%%%%%%%%%%%%%%%%%%%%%%%%%%%%%%%%%
\section{Introduction}
%%%%%%%%%%%%%%%%%%%%%%%%%%%%%%%%%%%%%%%%%%%%%%%%%%%%%%%%%%%%%%%

Suppose we are given a probabilistic model $P$ that captures a joint distribution over a stochastic sequence $X_1, X_2, \ldots, X_T$. We can use this probabilistic model to make sequential predictions: if we have observed a history $X_1=x_1$, \ldots, $X_t=x_t$ up to time~$t$, then we can predict the observation $X_{t+1}$ at time $t+1$ by conditioning the model on the past observations, that is,
\[
  P(X_{t+1}=x_{t+1} \mid x_1, \ldots, x_t).
\]
This prediction scheme works well under very broad conditions. In fact, it is optimal in expectation with respect to the log-loss; and if the sequence is governed by latent variables (which is the case, for instance, when the sequence is exchangeable) then the scheme corresponds to Bayesian prediction~\citep{cover1999elements}.

If we are interested in purposeful adaptive behavior (rather than mere prediction), it is natural to ask whether the above auto-regressive prediction scheme extends to the \emph{interactive setting}, where the model is used both for generating and predicting data (see e.g.~\citep{janner2021reinforcement, chen2021decision}). For instance assume we are given a probabilistic model for a stochastic sequence $A_1, O_1, A_2, O_2, \ldots, A_T, O_T$ describing sequential interactions between an expert and their environment, who issue the actions~$A_t$ and observations~$O_t$ respectively. Such a model could have been built from recorded demonstrations of an expert performing one or more tasks---say, a driver navigating a vehicle, a guided robot arm manipulating objects, or an interactive language model. First we remark that, as discussed before, we can use this model to passively predict the sequential expert-environment interactions. If we observe a sequence of past interactions $a_{1:t}, o_{1:t}$ between the expert and the environment, we can predict the expert's next action using the conditional probability
\[
  P(A_{t+1}=a_{t+1} \mid  a_{1:t}, o_{1:t}).
\]
The predictions made by this model will converge quickly even if we are initially uncertain about the task the expert is performing. 

But assume now that we want to use this model for \emph{imitation}. The rationale is to imitate the expert (or act as if one were the expert) by following the actions predicted by the model. The hope is that, as in the prediction case, if the sequence is governed by latent variables that characterize the task properties and the intentions of the expert, then the model will serve as an adaptive policy that converges quickly. More precisely, given past interactions~$a_{1:t}$ and~$o_{1:t}$, can we use this model for iteratively \emph{choosing} the next action~$A_{t+1}$ by sampling it from
\[
  a_{t+1} \sim P(A_{t+1} \mid a_{1:t}, o_{1:t}),
\]
where $a_{1:t}$ are the previous actions generated by the model and $o_{1:t}$ are the past observations produced by the environment? This is a simple question with far-reaching implications on foundation models~\citep{Bommasani:21foundation}, language agents~\citep{brown2020language}, behavior cloning~\citep{ho2016generative}, inverse reinforcement learning~\citep{ng2000algorithms, abbeel2004apprenticeship}, goal-/return- conditioned policies~\citep{veness2015compress, janner2021reinforcement, chen2021decision}), and so forth. For instance, it would allow the construction of highly-general adaptive agents by simply building a probabilistic model from observed demonstrations over multiple tasks and domains \emph{without} relying on expensive reward maximization procedures.

Unfortunately, the answer to this question is ``no''---at least not without assumptions (e.g.~that the task is a causally-sufficient Markov Decision Process and the expert policy is stationary) and not under the same general conditions under which passive sequence prediction works\footnote{It does work in some restricted settings discussed later.}. Unlike in control and reinforcement learning, here the actions are random variables, and as such embedded in a network of causal and probabilistic dependencies. In this report we will see that, while we can condition the model on observations as usual, actions (or action groups) must be treated as causal interventions. The reason is subtle: the model update triggered by the collected data differs depending upon whether the data was generated by the model itself (i.e.\ actions) or outside of it (i.e.\ observations), and mixing them up leads to wrong inferences. These take the form of \emph{self-delusions} where the model takes its own actions as evidence about the world (for instance, believing that ``Cheeseburgers are on the menu'' after sampling this sentence from the predictive distribution) due to the presence of confounding variables. This problem\footnote{The self-delusions discussed here are unrelated to the problem arising from observational confounding \citep{arjovsky2019invariant, zolna2019task, deletang2021causal}.} was first pointed out in~\citep{ortega2010bayesian, ortega2010minimum}, and related points were made in~\citep{rezende2020causally}.

%%%%%%%%%%%%%%%%%%%%%%%%%%%%%%%%%%%%%%%%%%%%%%%%%%%%%%%%%%%%%%%
\section{A minimal example}\label{sec:single}
%%%%%%%%%%%%%%%%%%%%%%%%%%%%%%%%%%%%%%%%%%%%%%%%%%%%%%%%%%%%%%%

To illustrate the self-delusion problem, we need an example with at least three random variables. For this, consider the following \emph{prize-or-frog} problem. There are two boxes (1 \& 2), one containing a prize and the other a frog respectively. The objective is to open the box containing the prize. 

\begin{figure}[t]
  \centering
  \def\svgwidth{0.9\textwidth}
  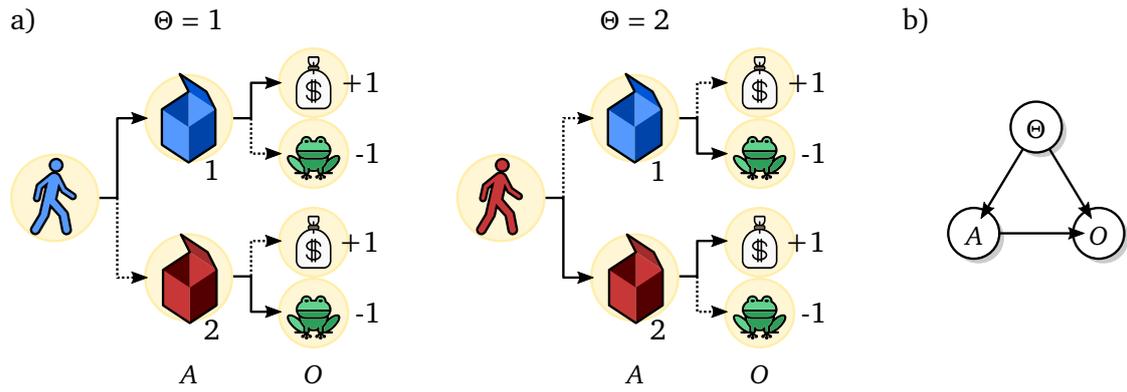
  \caption{The prize-or-frog problem. The objective is to choose the box containing the prize. Two problem instances exist: either the prize is in box~1 ($\Theta=1$) or box~2 ($\Theta=2$). There is also an expert who knows which box to open provided they are told the value of $\Theta$. Panel (a) depicts the two configurations. Solid transitions are deterministic (taken with probability one) and dotted ones are possible transitions that aren't taken. Panel (b) shows the causal Bayesian network of the problem.}
  \label{fig:boxes}
\end{figure}

Suppose we build a probabilistic model from data generated by an expert who opens the correct box when told where the prize is. For simplicity, we assume the two configurations are equiprobable. The full model is a joint distribution $P(\Theta, A, O)$ over the box configuration~$\Theta$, the chosen box~$A$ (1~or~2), and the observed content~$O$ ($\pm 1$ reward). See Figure~\ref{fig:boxes}a for a depiction. Formally, we have
\begin{small}
\[
 P(\Theta=\theta) = \frac{1}{2};
 \quad
 P(A=a \mid \Theta=\theta) =
 \begin{cases}
   1 & \text{if $a=\theta$,} \\
   0 & \text{otherwise;}
 \end{cases}
 \quad
 P(O=o \mid \Theta=\theta, A=a) =
 \begin{cases}
   1 & \text{if ($o=+1$ and $a=\theta$)} \\
     & \text{or ($o=-1$ and $a\neq\theta$),} \\
   0 & \text{otherwise.}
 \end{cases}
\]
\end{small}%
First we verify that the model works as intended for sequential prediction---that is, when the agent uses the model to sequentially predict the interactions between the expert and the task. We consider two cases: the fully observable case, and a partially observable where we do not see the task parameter.

\paragraph{Fully observable, passive.} 
This is a sequential prediction problem that proceeds in three steps. In each step, we predict the value of a random variable and then observe its outcome, following the order $\Theta$, $A$, and $O$. The predictions are made using the conditional probability distributions $P(\Theta)$, $P(A \mid \Theta)$, and $P(O \mid \Theta, A)$ respectively. Clearly, only the first prediction is uncertain---i.e.\ $P(\Theta=1) = P(\Theta=2) = \tfrac{1}{2}$; after $\Theta$ is revealed, $A$ and $O$ can be predicted with certainty.

\paragraph{Partially observable, passive.}
This time the configuration $\Theta$ is latent, and thus we are only required to predict the action~$A$ and the observation~$O$ using the marginal joint distribution
\[
  P(a, o) = \sum_\theta P(a, o \mid \theta) P(\theta).
\]
This marginal distribution can be derived as shown or learned directly from data, for instance (unintentionally) when~$\Theta$ is a hidden latent we are unaware of. Accordingly, we make our sequential predictions using the conditional probability distributions $P(A)$ and $P(O \mid A)$. The first prediction is uniform, that is
\begin{equation}\label{eq:first-action}
  P(a) = \sum_\theta P(a \mid \theta) P(\theta) = \frac{1}{2},
\end{equation}
which makes sense because we do not know the configuration $\Theta$. However, once the expert issues the action $A=a$, we can infer the unique value of $\Theta=\theta$ compatible with this choice. Given $A=a$, the subsequent observation~$O=+1$ occurs with certainty  because of the dependency between $\Theta$ and $A$. Mathematically, this is 
\[
  P(o \mid a) 
      = \sum_\theta P(o \mid \theta, a) P(\theta \mid a) =
      \begin{cases}
        1 & \text{if $o = +1$},\\
        0 & \text{if $o = -1$},
      \end{cases}
\]
where $P(\theta \mid a) = \delta[\theta = a]$ is the posterior probability of the configuration given the expert's action.

Now that we know the model works well for prediction, we will use the model in an interactive way for imitation, i.e.\ where we choose the action~$A$ following our sequential model instead of letting the expert choose it. Again we will consider both the fully and partially observable cases separately.

\paragraph{Fully observable, interactive.}
In the fully observable case we will first observe~$\Theta$, then choose~$A$, and finally observe~$O$. Clearly, once the value of~$\Theta$ is known, acting amounts to imitating the expert:
\[
  P(a \mid \theta) = 
    \begin{cases}
      1 & \text{if $a = \theta$,}\\
      0 & \text{otherwise.}
    \end{cases}
\]
After the action is chosen, the prediction of the outcome~$O$ follows as usual.

\paragraph{Partially observable, interactive---and the self-delusion problem.}
In this case we are required to first issue the action $A$ and then observe the outcome $O$, following the marginal distribution $P(A, O)$. Crucially, we do not observe $\Theta$.
  
Since we do not know the configuration of the task, the model is uncertain about which action to pick, suggesting $P(a)=\frac{1}{2}$ as calculated in \eqref{eq:first-action}. Sticking to the recommendation made by the model, we sample the action from $P(A)$. This immediately leads to a problem: \emph{whichever action we choose will convince ourselves of the box configuration}! That is,
\[
  P(\theta \mid a) =
    \begin{cases}
      1 & \text{if $\theta=a$,}\\
      0 & \text{otherwise,}
    \end{cases}
\]
because there is only one configuration that is consistent with the choice of the action. This \emph{delusion} will impact our inferences downstream: we now predict with certainty that we will obtain the prize $P(O=1 \mid a) = 1$ no matter which action we took, whereas clearly the correct inference is that we still don't know.

\begin{figure}[t]
  \centering
  \def\svgwidth{0.8\textwidth}
  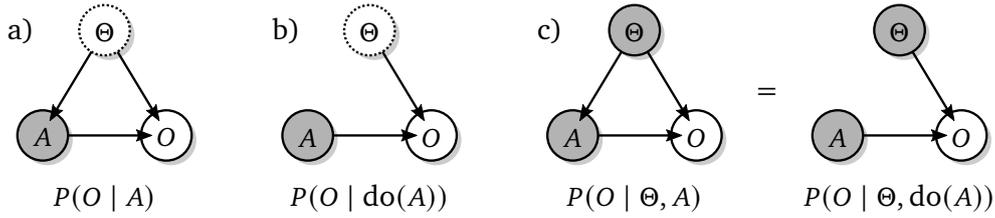
  \caption{The causal Bayesian network of the prize-or-frog problem, and the information flow resulting from different belief updates. Dotted outlines denote latent nodes, and shaded nodes mark the variables we condition on. Panel~(a) illustrates the self-delusion problem: conditioning on the self-generated action leads to wrong inferences about~$O$, since~$A$ and~$O$ are confounded by~$\Theta$. Panel~(b) shows that treating the self-generated action as a causal intervention circumvents the self-delusion because no information can flow backwards from~$A$ into~$\Theta$. Panel~(c) corresponds to the fully observable case. When~$\Theta$ is observed, then conditioning or intervening on the self-generated action leads to the same prediction over~$O$.}
  \label{fig:boxes-dag}
\end{figure}

\paragraph{Addressing the delusion.} How did this delusion happen and does one circumvent it? The problem lies in the causal structure: $\Theta$ causally precedes $A$ and $O$, and $A$ precedes $O$ (Figure~\ref{fig:boxes}b). Crucially, the causal structure implies that knowing the configuration~$\Theta$ is a precondition for choosing the action~$A$, just as the expert did. But when we choose the action using~$P(A)$ without knowing~$\Theta$ we are violating the causal requisite. This is why conditioning on the generated action will not only provide evidence about the causal future, but erroneously also about the causal past (Figure~\ref{fig:boxes-dag}a). In short, $\Theta$ is a \emph{confounder} \citep{pearl2009causality}.

To avoid this mistake, one must incorporate the causal independence constraint introduced by the choice of the action~$A$ without the knowledge of~$\Theta$, formally written as $\Theta \indep A$. Mathematically, this is done by treating the action as a causal intervention on~$P$ using the \emph{do-operator}, resulting in the posterior distribution $P(\Theta \mid \DO(A))$. A calculation shows that intervening on $A$ does not change the beliefs about $\Theta$:
\[
  P(\theta \mid \DO(a))
  = \DO(a) \biggl\{ 
  \frac{ P(a \mid \theta) P(\theta) }
  { \sum_{\theta'} P(a \mid \theta') P(\theta') }
  \biggr\}
  = \frac{ P(\DO(a) \mid \theta) P(\theta) }
  { \sum_{\theta'} P(\DO(a) \mid \theta') P(\theta') }
  = P(\theta),
\]
where we used Bayes' rule and then the fact that $P(\DO(a) \mid \theta) = Q(a \mid \theta) = Q(a)$ for some $Q$ where $\Theta \indep A$. Unlike conditioning, intervening on the generated action breaks the causal link between~$\Theta$ and~$A$, which in turn correctly provides evidence about the causal future, but not on the past~(Figure~\ref{fig:boxes-dag}b). Consequently, the prediction of the observation $O$ is given by the posterior predictive $P(O \mid \DO(A))$ which can be obtained via the back-door adjustment formula:
\[
  P(o \mid \DO(a)) 
  = \sum_\theta P(o \mid \theta, \DO(a)) P(\theta \mid \DO(a))
  = \sum_\theta P(o \mid \theta, a) P(\theta)
  = \frac{1}{2}.
\]
The result makes sense: now, selecting a box at random using the model does not provide evidence about the configuration, which in turn leaves us uncertain about the outcome. 

\bigskip

\begin{rem}
We stress that, unlike in standard conditioning, to evaluate an intervention we need to know the causal structure. For instance, had the graph been the same except that~$A$ causally precedes~$\Theta$, then the conditional probability $P(\theta \mid \DO(a))$ would have been
\[
  P(\theta \mid \DO(a))
  = \DO(a) \biggl\{ 
  \frac{ P(a) P(\theta \mid a) }
  { \sum_{\theta'} P(a) P(\theta' \mid a) }
  \biggr\}
  = \frac{ P(\DO(a)) P(\theta \mid \DO(a)) }
  { \sum_{\theta'} P(\DO(a)) P(\theta' \mid \DO(a)) }
  = P(\theta \mid a) = \delta_a^\theta.
\]
In this case choosing the action does fix the box configuration, consistent with the new causal story.
\end{rem}

\paragraph{Back to fully observable, interactive.}
Having seen that actions are causal interventions one might ask: why did we not intervene in the fully observable case? The answer is: we did. The difference is that in the fully observable case, conditioning and intervening leads to the same posterior distribution: 
\[
  P(O \mid \Theta, A) = P(O \mid \Theta, \DO(A)).
\]
This is seen by noticing that conditioning on~$A$ only produces a flow of information that travels downstream. The upward flow of information is blocked, since we have already conditioned on $\Theta$ (Figure~\ref{fig:boxes-dag}c).

\paragraph{Summary.}
When using a probabilistic model in an interactive setting, actions (i.e.\ data generated by the model) are causal interventions, whilst observations (i.e.\ data generated externally) are standard Bayesian conditions. This prevents incorrect inferences (=delusions) via hidden confounders. Hence, one cannot learn from one's own actions, but only from their effects. Furthermore, to evaluate an intervention we need to know the underlying causal structure.

\bigskip

\begin{rem}
In our example situation we used a random configuration, but otherwise the expert choices and outcomes are deterministic. However, the reader can easily verify that the problem and the solution hold in the general case when the expert choices and the outcomes are stochastic.
\end{rem}

\begin{rem}
The source of the delusion can also be seen by comparing the two posterior beliefs $P(\theta \mid a, o)$ and $P(\theta \mid \DO(a), o)$ over the task parameter, that is, where the past action is treated as a standard condition and a causal intervention respectively (differences are highlighted):
\begin{align}\label{eq:delusion}
  P(\theta \mid {\color{red} a}, o)
  &= \frac{ P(\theta) {\color{red}P(a \mid \theta)} P(o \mid \theta, a) }{
  \sum_{\theta'} P(\theta') {\color{red}P(a \mid \theta')} P(o \mid \theta', a)}
  &
  P(\theta \mid {\color{red} \DO(a)}, o)
  &= \frac{ P(\theta) P(o \mid \theta, a) }{
  \sum_{\theta'} P(\theta') P(o \mid \theta', a)}.
\end{align}
Here we can clearly see that the intervention causes the dismissal of the evidence $P(a \mid \theta)$ produced by the self-generated action~$A=a$.
\end{rem}

\begin{rem}
In control and RL, the controller/agent is never included into the probabilistic model, but regarded as an external process. Actions are not treated as random variables but as indexes over families of distributions. Hence, when the control/RL practitioner ``conditions'' on an action, technically they choose a distribution from a family parameterized by the actions, which results in the (correct) second posterior belief formula in~\eqref{eq:delusion}.
\end{rem}

\begin{rem}
Sampling an action from the posterior predictive $P(a)$ amounts to sampling from
\[
  P(a) = \sum_\theta P(a \mid \theta) P(\theta).
\]
The sum (or integral, if $\Theta$ were continuous) can render the direct calculation of $P(a)$ intractable. An equivalent and  algorithmically easier way to obtain an action using Monte-Carlo simulation is to first sample $\theta \sim P(\theta)$ and then sample the action from the expert policy $a \sim P(a \mid \theta)$. Note that this is Thompson Sampling---not as a heuristic, but derived from first principles \citep{ortega2010bayesian, ortega2010minimum}.
\end{rem}

\begin{rem}
This is also the reason why goal-conditioned planning does not work in the general case. Suppose the interactions are generated from the process $\Theta, A, O, G$, where $G$ is a random variable characterizing the final state that depends causally on~$\Theta$. This could e.g.\ be an explicit final state or a function computed from the interactions, such as the return. In goal-conditioned planning, one conditions on the desired goal $G=g$ and then acts according to the posterior predictive, i.e.\ $P(A \mid G=g)$. The problem is that the choice of the goal $G=g$ is an action, and conditioning on it will reveal false information about the latent task parameter~$\Theta$, which is delusion.
\end{rem}

\begin{rem}
The breakdown in the partially observable case relied on the presence of an unobservable latent factor, which in general cannot be avoided. However, often there are domains where the observation is so rich that it (implicitly) contains the task identifier. This essentially brings the problem back to the fully observable case, limiting the impact of the delusion.
\end{rem}

\begin{rem}
The role played by interventions in our discussion has a parallel in the brain sciences. The central motor system, when issuing a motor signal, generates an efferent copy which is sent to the sensory system. This enables the sensory system to take into account what the intended actions are, so that these are not mistaken for environmental actions. This mechanism is hypothesized to be the reason why one can't tickle oneself \citep{blakemore2000can}.
\end{rem}

%%%%%%%%%%%%%%%%%%%%%%%%%%%%%%%%%%%%%%%%%%%%%%%%%%%%%%%%%%%%%%%
\section{The sequential case}\label{sec:sequential}
%%%%%%%%%%%%%%%%%%%%%%%%%%%%%%%%%%%%%%%%%%%%%%%%%%%%%%%%%%%%%%%

A straightforward way of extending the previous example to sequential decisions is by formalizing the interactions as a sequential game because they reflect the causal dependencies. We will do this in two steps: first we formulate the idealized, causally sufficient game which includes all the task parameters an expert needs in order to make their decisions, and then we use it to derive an adaptive policy for when these parameters are latent. The fact that this actually works is non-trivial.

\begin{figure}[t]
  \centering
  \def\svgwidth{\textwidth}
  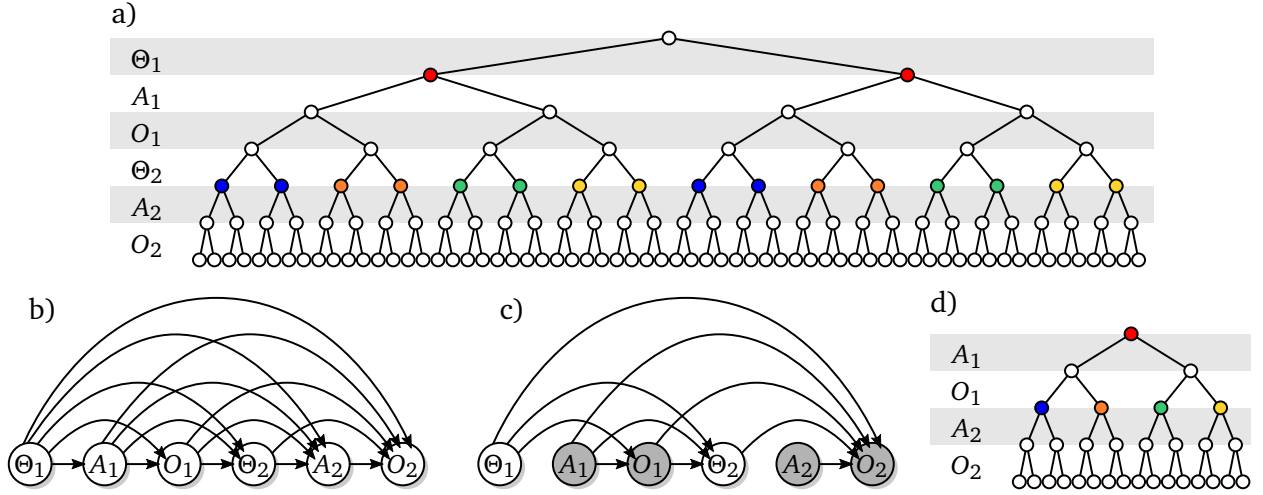
  \caption{Stochastic process over the binary random variables $\Theta_1, A_1, O_1, \Theta_2, A_2, O_2$ representing interactions between Nature (choosing $\Theta$'s and $O$'s) and the agent (choosing $A$'s). The interactions can be represented as a game in extensive form (Panel a) or as a causal Bayesian network (Panel~b). In the partially observable case the $\Theta$'s are hidden moves by Nature, which create information sets (=collection of indistinguishable states) for the agent. These are shown explicitly in~(a) as nodes sharing the color which reduce the game to~(d). Panel (c) shows the graph resulting from treating actions as interventions. Panel (d) shows the game representing the marginal interaction process that the agent experiences when Nature makes hidden moves. This game is causally insufficient for evaluating the effects of interventional actions.} 
  \label{fig:sequential-game}
\end{figure}

We start by considering a stochastic process~$P$ with~$T$ rounds over the random variable triplets
\[
  \Theta_1, A_1, O_1, \Theta_2, A_2, O_2, \ldots, \Theta_T, A_T, O_T,
\]
where~the $\Theta_t$ and $O_t$ are moves taken by Nature, and the~$A_t$ are moves taken by an expert. Here we chose this triplet structure only for convenience: one can easily envision games with other structures better suited for the desired application domain. For instance, a natural language domain might require hierarchically structured variables (say, topic, sentence, and word markers). Figure~\ref{fig:sequential-game}a shows an example with binary random variables and $T=2$ rounds and Figure~\ref{fig:sequential-game}b shows its associated causal Bayesian network. As in the prize-or-frog game, we interpret the $\Theta$'s as task/expert-intention parameters; only here they are generated as the process unfolds. The process represents a game with complete information, where every move is observable by both parties. That is, in every time step~$t+1$, the expert sees all the past moves and chooses an action from the conditional distribution (i.e.\ the expert policy)
\[
  P(A_{t+1} \mid \theta_{1:t+1}, a_{1:t}, o_{1:t}).
\]
Variations of such a game could represent, for instance, an agent playing a multi-armed bandit with known reward distributions; an agent controlling a known MDP; or a card player who sees the hands of their opponents (the opponent players were folded into a single environment).

Now let us assume we use this model to mimic the interactions, but without seeing the task parameters. The interaction process is thus the marginal process over the random variables
\[
  A_1, O_1, A_2, O_2, \ldots, A_T, O_T.
\]
In the language of game theory, we have turned the original, perfect information game into a game with imperfect information. In this more challenging game, Nature can make hidden moves, choosing the task parameters~$\Theta_t$ secretly. From game theory, we also know that hidden moves spawn \emph{information sets}, i.e.\ game states that are indistinguishable (highlighted in Figure~\ref{fig:sequential-game}a) and hence do not admit the same state-specific actions that the expert would choose \citep{vonNeumann1947theory, osborne1994course}.

Given the above limitations, how do we choose our next action? As we have seen in the prize-or-frog example, we cannot sample actions from $P(A_{t+1} \mid a_{1:t}, o_{1:t})$ because conditioning on past actions leads to delusions. Instead, we must choose our next action using
\begin{equation}\label{eq:action-pred}
  P(A_{t+1} \mid \DO(a_{1:t}), o_{1:t})
\end{equation}
where past actions are incorporated as causal interventions and past observations as standard conditions. This leads to inferences that are consistent with the causal constraints imposed by the information sets. Compare the causal Bayesian network before and after the interactions in Figures~\ref{fig:sequential-game}b \&~c respectively. In addition, the resulting policy is adaptive---essentially, it is Thompson sampling---and will converge to the expert's policy under broad conditions \citep{leike2016thompson, osband2017posterior, ortega2010bayesian}.

If instead we had started directly from the marginal process over the interactions without knowing the underlying causal dependencies on the confounders~$\Theta_t$ (see Figure~\ref{fig:sequential-game}d), then we would have been incapable of evaluating the correct result of interventions. This is why possessing a causally sufficient model is of paramount importance.

\paragraph{Summary.}
If we want to imitate an expert who interacts with a task class, we can do so by constructing a causally sufficient model of their sequential interactions, and then sample our actions from the conditional~\eqref{eq:action-pred} which treats past actions as interventions. In short, imitation requires knowing the reasons behind actions. This will produce an adaptive policy that converges to the expert's policy under broad conditions. 

\bigskip

\begin{rem}
Interestingly, by introducing the notion of information sets for describing hidden moves, Von~Neumann shows to have been acutely aware of the associated causal problem already in 1944 \citep[Chapter~7]{vonNeumann1947theory}.
\end{rem}

\begin{rem}
While the conditional distribution~\eqref{eq:action-pred} looks simple, it is worth expressing it in terms of conditional probabilities that are free from interventions. For completeness, here we provide its recursive definition. The posterior predictive over the next action is
\[
  P(a_{t+1} \mid \DO(a_{1:t}), o_{1:t})
  = \sum_{\theta_{1:t+1}} P(a_{t+1} \mid \theta_{1:t+1}, a_{1:t}, o_{1:t}) \, P(\theta_{1:t+1} \mid \DO(a_{1:t}), o_{1:t}),
\]
i.e.\ a weighted superposition of expert actions weighted by the posterior probabilities of the task parameters. The posterior $P(\theta_{1:t+1} \mid \DO(a_{1:t}), o_{1:t})$ in turn is given by the recursive expression
\[
  P(\theta_{1:t+1} \mid \DO(a_{1:t}), o_{1:t})
  \propto 
    P(\theta_{t+1} \mid \theta_{1:t}, a_{1:t}, o_{1:t})
    \, P(o_t \mid \theta_{1:t}, a_{1:t}, o_{1:t-1})
    \, P(\theta_{1:t} \mid \DO(a_{1:t-1}), o_{1:t-1}),
\]
That is, proportional to the product of the probability of choosing the last task parameter, the likelihood of the observation, and the prior probability of the remaining task parameters. The likelihood of the action is absent from the product because it was treated as an intervention. This derivation makes use of standard probability and the rules of causal calculus. It is easy to see that if the recursive equation is ``unrolled'', the result will be entirely free of expressions containing interventions. 
\end{rem}

\begin{rem}[Multi-armed bandits]
Learning how to play a multi-armed bandit is the quintessential introductory reinforcement learning problem \citep{sutton2018reinforcement}, exemplifying the exploration-exploitation dilemma an agent must face when dealing with uncertainty. We built a model of interactions between a noisy expert and a multi-armed bandit with $5$~arms where each arm delivers a Bernoulli-distributed reward. The interaction process is $\Theta, A_1, O_1, A_2, O_2, \ldots$ where $\Theta$ is the hidden choice of the best arm ($\theta \in \{1, 2, 3, 4, 5\}$), and the $A_t$ and $O_t$ are the expert actions and the bandit rewards respectively. The causal probabilistic model is specified by the following mechanisms,
\[
  P(\Theta=\theta) = \frac{1}{5};\quad
  P(A_t=a_t \mid \theta) = 
  \begin{cases}
    0.6 & \text{if $a_t=\theta$,} \\
    0.1 & \text{otherwise;}
  \end{cases}\quad
  P(O_t=1 \mid \theta, a_t) =
  \begin{cases}
    0.75 & \text{if $a_t=\theta$,} \\
    0.25 & \text{otherwise.}
  \end{cases}
\]
Since these are the mechanisms, the $A_t$ are independent of the past given $\Theta$, and so are the $O_t$ given the previous action $A_t$ and the parameter $\Theta$. The table below shows the result of using this model for imitating the expert during twenty rounds without knowing~$\Theta$. It compares two cases: when past actions are treated as standard conditions and as causal interventions with three trajectories each. For simplicity, the best arm was labelled as arm~1. 
\begin{center}
\footnotesize
\begin{tabular}{clccl}
\toprule
\multicolumn{2}{c}{Conditioning: $a_{t+1} \sim P(a_{t+1} \mid a_{1:t}, o_{1:t})$} && \multicolumn{2}{c}{Intervening: $a_{t+1} \sim P(a_{t+1} \mid \DO(a_{1:t}), o_{1:t})$} \\
\midrule 
Actions: & 3 4 3 3 3~~3 3 3 3 3~~3 3 3 3 3~~3 3 3 3 3 
&& Actions: & 1 4 4 3 1~~1 1 5 3 1~~1 1 4 1 1~~1 1 1 1 1 \\
Rewards: & 1 0 0 0 0~~0 0 1 0 0~~0 0 1 0 0~~0 1 0 1 0 
&& Rewards: & 0 0 0 1 1~~1 0 0 0 1~~1 1 0 1 1~~1 1 1 0 1 \\
\\
Actions: & 2 1 2 1 2~~1 1 1 1 1~~1 1 1 1 1~~1 1 1 1 1 
&& Actions: & 5 4 5 3 4~~5 3 3 2 3~~1 1 1 5 5~~1 1 1 1 1 \\
Rewards: & 0 1 0 1 0~~1 1 1 1 0~~1 1 1 1 1~~1 1 0 1 1 
&& Rewards: & 1 0 0 1 0~~0 0 1 0 0~~1 0 1 1 0~~1 1 1 0 1 \\
\\
Actions: & 5 5 2 2 2~~2 5 2 4 1~~2 2 2 2 2~~2 2 2 2 5 
&& Actions: & 2 1 1 1 5~~4 5 4 1 4~~5 1 1 1 1~~1 4 1 1 1 \\
Rewards: & 0 0 1 0 1~~0 0 0 1 1~~1 1 0 0 1~~0 0 0 0 0 
&& Rewards: & 0 1 1 0 0~~0 0 0 0 0~~1 1 1 1 1~~0 0 1 1 1 \\
\bottomrule
\end{tabular}
\end{center}
The two resulting strategies differ significantly. Treating actions as conditions leads to self-delusion---evidenced by a tendency to repeating previously chosen actions, often converging to the wrong arm. In contrast, treating actions as interventions leads to a richer exploratory strategy which identifies the best arm.
\end{rem}

\begin{rem}[Language models]
In language modelling, the actions and observations are both language tokens, such as words. The web is full of the actions (text) produced by many other agents, mostly people, but recently machines too, such as GPT3 \cite{brown2020language}. Language models $P(x_t \mid x_{1:t-1})$ are often pre-trained with self-supervised learning techniques.
These pre-trained models are agents that can generate actions by conditioning on previous actions. Their environment is an API that allows for human experts to prompt the agents with text. 

Consider a pre-trained language model whose job is to predict the fourth word given a sequence of three words proposed by us (experts). Without any loss of generality, we again introduce the variable $\theta$ to capture information that is available to the experts, but not to the agent. This information could represent the intention, emotional state or any other information about the experts that the language model has no access to via the observations. If we knew the posterior distribution of $\theta$, we could easily marginalise it out using the rules of probability, and predict the fourth word as follows:
\begin{equation}
P(x_4 \mid x_1, x_2, x_3) = 
\int P(x_4 \mid \theta, x_1, x_2, x_3) 
\, P(\theta \mid x_1, x_2, x_3) \, d\theta,
\label{eq:pred1}
\end{equation}
where the posterior distribution $P(\theta \mid x_1, x_2, x_3)$, representing the model's beliefs about $\theta$, is given by
\begin{equation}
P(\theta \mid x_1, x_2, x_3) 
  \propto P(x_3 \mid \theta, x_1, x_2)
    \, P(x_2 \mid \theta, x_1) \, P(x_1 \mid \theta) \, P(\theta). 
\label{eq:post1}
\end{equation}
Note that in practice we don't know the shape of $\theta$ and indeed it may even be an unknown unknown. Therefore we don't compute the prediction using the right hand side of equation~(\ref{eq:pred1}), but instead train the predictive model on the left hand side of this equation. 

The problem arises when we consider interaction between the pre-trained language model and the environment. Suppose we use a language model API to enter the first word $x_1$, but this time let the model use its own generated second word $x_2$, and then force the model to use our third word $x_3$. We then try to predict the fourth word as before. That is, while $x_1$ and $x_3$ are expert data, $x_2$ is an intervention produced by the language agent. Hence, by the arguments introduced earlier in the minimal example section, we must adopt the following model to avoid delusions:
\begin{equation}
P(x_4 \mid x_1,\DO(x_2),x_3) = 
  \int P(x_4 \mid \theta,x_1,x_2,x_3) 
    \, P(\theta \mid x_1,\DO(x_2),x_3) \, d\theta
\end{equation} 
The posterior distribution $P(\theta \mid x_1,\DO(x_2),x_3)$, with $x_2$ being treated as an intervention, is given by
\begin{equation}
P(\theta \mid x_1,\DO(x_2),x_3) 
  \propto P(x_3 \mid \theta,x_1,x_2)
    \, P(x_1 \mid \theta) \, P(\theta) 
\label{eq:post2}
\end{equation}
Note that $P(x_2 \mid \theta,x_1)$ appears in equation~(\ref{eq:post1}) but not in equation~(\ref{eq:post2}) because in this latter equation what we have is $P(\DO(x_2) \mid \theta,x_1)$ and this term cancels out in the expression for the posterior distribution as we discussed earlier. This can be understood as an application of the back-door criterion to the causal graph, whereby we delete all links pointing to $x_2$. That is, $x_2$ is fixed, and not a random variable. Hence,
\begin{equation}
P(x_4 \mid x_1,x_2,x_3) \neq P(x_4 \mid x_1,\DO(x_2),x_3). 
\end{equation}
If the model acted in the past, its actions must be treated as causal interventions and not as observations. In the extreme case imagine that the language model generates a lot of text and that this text is added to the dataset, say a web corpus. Then relearning from this dataset will only confirm the model's biases, that is, its delusions. 

How important is this in practice? If the observations are very informative about the latent information $\theta$, then the difference between the two predictive models, described above, will be small. However, when the observations are not informative about $\theta$, or ambiguous, and the agent cannot collect data online to reduce its ignorance, the two predictive models can be very different, giving rise to absurd dialogues as the one in the introduction of this paper. 
\end{rem}

%%%%%%%%%%%%%%%%%%%%%%%%%%%%%%%%%%%%%%%%%%%%%%%%%%%%%%%%%%%%%%%
\section{Sequential models built from data}
%%%%%%%%%%%%%%%%%%%%%%%%%%%%%%%%%%%%%%%%%%%%%%%%%%%%%%%%%%%%%%%

So far we have assumed that we are in possession of a complete causal-probabilistic model over interaction sequences on which we can arbitrarily operate by marginalizing, conditioning, and intervening on any desired random variable. We now turn our attention to learning such models from data through regression, where the models typically have rigid function signatures. For instance, say we learn a function~$f$ that computes the probabilities of a Markov transition kernel 
\[
  P(X_{t+1}=x_{t+1} \mid X_t=x_t) = f(x_{t+1}, x_t).
\]
Using $f$, we cannot compute $P(X_{t+1} \mid \DO(x_t))$. This is the type of challenge we'll face next.

\subsection{Meta-learning and ``counterfactual teaching''}

We can learn a sequential model for control using memory-based meta-learning \citep{duan2016rl, wang2016learning} (as was done to train e.g.\ GPT-3 \citep{brown2020language}). If we have a collection of tasks and an expert, and a prior distribution over task parameters, then we can meta-train an agent system with memory to learn an adaptive policy over the task class. An agent with memory could be e.g.\ a recurrent neural architecture based on LSTM cells or a higher-order Markov model based on transformers. Once the agent has been meta-trained, it can then be deployed at test time with fixed parameters; importantly, the resulting agent will follow an adaptive policy.

\paragraph{Setup.}
For simplicity here we restrict ourselves to interaction processes with one initial latent parameter, i.e.\ $\Theta, A_1, O_1, \ldots, A_T, O_T$. Formally, we need the following ingredients: A \emph{prior distribution over task parameters}~$Q(\Theta)$; a \emph{collection of tasks}, where each task is represented by a conditional probability distribution~$Q(O \mid \Theta, W)$ over the next observation~$O$ given the task parameter~$\Theta$ and the current memory state~$W$; an \emph{expert}, represented by a conditional probability distribution~$Q(A \mid \Theta, E)$ over the next ideal action $A$ given the task parameter~$\Theta$ and the current memory state~$E$; and an \emph{agent}, represented by learnable conditional probability distributions~$P(A \mid M)$ and~$P(O \mid M)$ predicting the next action or observation given a memory state~$M$. In all of the three, the memory states $W$, $E$ and $M$ act as the sufficient statistics of the past interactions. 

Without loss of generality, we  can implement our agent using four trainable functions,
\begin{align*}
  P(A \mid M=m) &= f_A(m) &
  P(M' \mid M=m, A=a) &= g_A(a, m) \\
  P(O \mid M=m) &= f_O(m) &
  P(M' \mid M=m, O=o) &= g_O(o, m)
\end{align*}
where~$f_A$ and~$f_O$ implement the policy and the prediction, and where~$g_A$ and~$g_O$ are their associated memory transition kernels. 

\begin{figure}[t]
  \centering
  \def\svgwidth{\textwidth}
  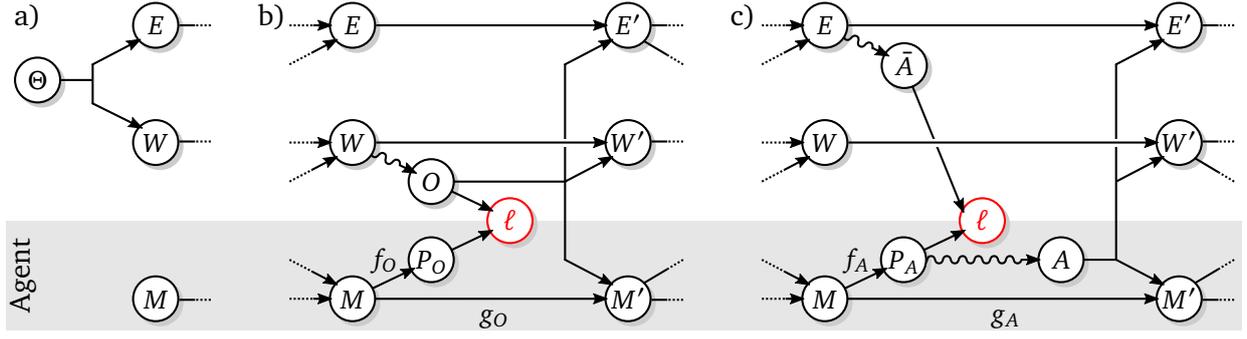
  \caption{Computation graph for meta-training. Squiggly arrows ($\rightsquigarrow$) denote stochastic functions. a)~Initialization step. $\Theta$ is the randomly drawn task parameter, which is shown to the expert. Crucially, $\Theta$ is not shown to the agent. $E$, $W$, and $M$ are the expert's, the task's, and the agent's memory states respectively. b)~Observation step with factual teaching. The agent predicts the next observation using a distribution~$P_O$. The task responds stochastically with a factual observation~$O$ which generates a log-loss~$\ell$ and updates the memory states. c)~Action step with counterfactual teaching. The agent predicts the expert's action using a distribution~$P_A$ and samples an action~$A$ from it, which updates the memory states. Crucially, this stochastically generated action must not propagate gradients when trained end-to-end using backpropagation---if necessary, a stop-gradient must be introduced to model the intervention. The expert then stochastically generates an ideal action $\bar{A}$ which generates a log-loss~$\ell$ for the agent's prediction. ``Counterfactual teaching'' refers to the property that the agent doesn't get penalized for the action it takes, but for its probability of taking the action the expert would have taken.}
  \label{fig:computation-graph}
\end{figure}

\paragraph{Training.}
The challenge is to meta-train the agent so that we address the following two problems:
\begin{itemize}
    \item[a)] the observation predictions regress the Bayesian posterior predictive;
    \item[b)] and the action probabilities regress the posterior predictive, but respecting the causal constraints before and after choosing the action.
\end{itemize}
Part~(a) is solved by minimizing the log-loss of the predictions in the standard way. We call this \emph{factual teaching}. To address~(b), the agent first predicts the expert's action and then samples its own action from it. Subsequently the expert reveals their action, inducing a log-loss penalty for the agent. We call this scheme \emph{counterfactual teaching}. This training setup makes sure that the resulting agent, at deployment time, imitates the expert while treating past actions as interventions (see Appendix~\ref{sec:fcf-teaching}). The computation graph is shown in Figure~\ref{fig:computation-graph}.

It is worth being clear about what this achieves. If the agent is trained using factual/counterfactual teaching to regress the observation/action probabilities, then upon convergence the functions~$f_A$ and~$f_O$ will approximate\footnote{Assuming the function classes are rich enough.}
\begin{align*}
  f_A(m_t) &\approx Q(A_{t+1} \mid \DO(a)_{1:t}, o_{1:t}) \\
  \text{and}\quad
  f_O(m'_t) &\approx Q(O_{t+1} \mid \DO(a)_{1:t+1}, o_{1:t}),
\end{align*}
where~$m_t$, and $m'_t$ are the sufficient statistics for the interaction histories and where all the past actions---and only those---have been intervened on. This avoids the delusions pointed out in Sections~\ref{sec:single}~\&~\ref{sec:sequential}. Other probabilities, such as
\[
Q(A_{t+1}=a_{t+1} \mid a_{1:t}, o_{1:t})
\quad \text{or} \quad
Q(A_{t+1}=a_{t+1} \mid a_{1:t}, \DO(o)_{1:t}),
\]
where we condition on actions or where we intervene on the observations respectively, are not captured. 

\paragraph{Summary.}
When training a function approximator to learn an adaptive policy, the causal distinction between actions and observations translates into using counterfactual and factual teaching signals respectively. This ensures that predictions are amortized using the correct weighting of past histories that mix conditioning and intervening.

\bigskip

\begin{rem}
Rather than using an architecture with explicit internal memory, alternatively we can also train a single function implementing a sequence predictor
\[
  P(X_{t+1} \mid X_1=x_1, \ldots, X_t=x_t) = f([x_1, x_2, \ldots, x_t]).
\]
In this approach the architecture does not distinguish between actions and observations, treating them uniformly as symbols. Since the entire past is provided as prediction context, no explicit memory is required because the past is trivially a sufficient statistic of itself. Even when the context is truncated to the most recent~$K$ symbols (=$K$-order Markov) it could still provide a good approximation to the ideal sufficient statistic (provided~$K$ is sufficiently large). 
\end{rem}

\begin{rem}
For all practical purposes, a given symbol~$x$ and its intervened version~$\DO(x)$ are best seen as entirely different symbols. For instance,
\[
  P(X_4 \mid \DO(x_1), x_2, x_3),
  \qquad
  P(X_4 \mid x_1, \DO(x_2), x_3),
  \qquad\text{and}\qquad
  P(X_4 \mid x_1, \DO(x_2), \DO(x_3)),
\]
are three different conditional probability distributions because they differ in their intervened variables: each must be meta-learned separately, carefully switching between factual and counterfactual teaching signals. Thus in the previous example, assuming that $X_1$, $X_2$, and $X_3$ are binary, then there are not $2^3=8$, but $4^3=64$ different pasts of length 3, because each random variable has~2 standard and~2 intervened possible values. 
\end{rem}

\begin{rem}\label{rem:distinguish}
In the example given we explicitly distinguished between the task and the expert as providers of the factual and counterfactual teaching signals respectively. However, in many cases one cannot or does not want to draw this distinction. A timely example is given by interactive language models. Tokens generated by the model are actions to be treated as causal interventions, while tokens provided by the user are standard observations. Since we cannot tell ahead of time which party will produce which part of speech, it would be necessary to amortize each token both in its standard and intervened form using factual and counterfactual teaching signals. 
\end{rem}

\begin{rem}\label{rem:delude}
If we do not distinguish between actions and observations, then a language model will delude itself. In the opening epigraph of this report, the sentences generated by the model were incorporated into the prediction context as standard conditions.
\end{rem}

\begin{rem}
The requirement of soliciting an expert's action online from a trajectory generated by an agent for counterfactual teaching is similar to the Dagger algorithm \citep{ross2011reduction} where an expert provides supervising signals for imitation learning in states reached by a student agent. However, they come from different motivations. Dagger is designed to reduce the extrapolation error due to distribution shift at test time, and doesn't consider the existence of latent parameters. The benefit of Dagger over vanilla imitation learning diminishes if a student can predict the expert's action distribution exactly. In contrast, it becomes necessary to request counterfactual learning signals from an expert in this study at the presence of latent variables. The bias always exists in this case no matter how expressive the agent model is and how much data is available.
\end{rem}

\begin{rem}\label{rem:onlydoneeded}
For the purpose of learning an adaptive policy, it appears that all we want is estimate the action distribution $P(A_{t+1} \mid \DO(a)_{1:t}, o_{1:t})$. Therefore, training our model on the loss on $f_A(m)$ might be enough, although the loss defined for observation prediction could be useful as an auxiliary loss.
\end{rem}

\subsection{Offline Adaptation and Control}

Now our goal is to train an agent to imitate an expert who is not available. Instead, we only have a dataset containing demonstrations, i.e.\ trajectories of interactions between the expert and the tasks, from which we want to build our (causal-probabilistic) sequential model. This limitation does not impact learning a sequential predictor, but it does affect the problem of learning an adaptive policy in a fundamental way. Currently this is an open problem, and we will limit ourselves to merely pointing out the main difficulty.

As before we assume the interaction process is $\Theta, A_1, O_1, \ldots, A_T, O_T$, but now we are only provided with sampled trajectories of the form $\tau = a_1, o_1, \ldots, a_T, o_T$ where the task parameter~$\theta$ is latent. Meta-training an agent for sequential prediction is easy: we select a trajectory at random, and then we sequentially provide factual teaching signals for the agent's predictions. However, in the interactive case, if we do not know the task  parameter~$\theta$, then we cannot supply counterfactual teaching signals. 
To see this, imagine we attempt to meta-train following the approach of the prediction case, and sample a trajectory $\tau = \bar{a}_1, o_1, \ldots, \bar{a}_T, o_T$ from the dataset. Here the $\bar{a}_t$ are the expert actions. The agent starts by forecasting the first action using~$P(A_1)$. To implement the counterfactual teaching signal, we penalize the prediction using the expert's action as $\ell = -\log P(A_1=\bar{a}_1)$ and let the agent pick its own action $a_1 \sim P(A_1)$. If the two actions differ, i.e.\ $a_1 \neq \bar{a}_1$, then the agent's choice is incompatible with our sampled trajectory~$\tau$ and hence we cannot use it to continue the sequence. This problem cannot be circumvented by replacing~$\tau$ with another sequence~$\tau'$ from the dataset that starts with~$a_1$, because $\tau'$ might not have been generated by a task-expert combination having the same latent parameter~$\theta$ as the original trajectory~$\tau$.

This is not a problem of data sparsity: even if the dataset is rich enough to contain every possible interaction sequence (with different multiplicities) the above problem persists because the data has been collected under the presence of the confounding factor~$\Theta$.
To address the problem, it is necessary to incorporate assumptions (e.g.\ a set of hypotheses) about the confounding variable, but currently it is unclear how to incorporate this seamlessly into a meta-learning scheme.

\paragraph{Summary.}
In general, we cannot meta-train an agent only from expert demonstrations to imitate the expert at deployment time, because said demonstrations could depend on unknown confounding variables. To train an agent using expert demonstrations, it is necessary to induce causal models which propose the necessary confounders for explaining the data.

%%%%%%%%%%%%%%%%%%%%%%%%%%%%%%%%%%%%%%%%%%%%%%%%%%%%%%%%%%%%%%%
\section{Conclusion}
%%%%%%%%%%%%%%%%%%%%%%%%%%%%%%%%%%%%%%%%%%%%%%%%%%%%%%%%%%%%%%%

In this report we have explored the use of sequence models for control. This was motivated by the recent enormous success of large language models. We have shown that a naive use, where actions are chosen according to the predicted next action conditioned on the past, does not work in general because it creates delusions, i.e.\ situations in which the agent mistakes its own actions for evidence about the task. These delusions occur whenever there are confounding factors that obscure the cause-effect relationships between actions and observations.

To overcome this, actions must be treated as causal interventions after being generated. Doing so introduces the correct causal constraints so that the agent can only learn about the task through the effects of its actions. However, performing interventions requires having access to the causal model. We then explored meta-learning a sequence model suitable for adaptive imitation. The main challenge here is to make sure that actions and observations are regressed as interventions and conditions respectively, and we have shown that this can be achieved via counterfactual and factual teaching signals respectively. 

Finally, if the sequence does not possess any latent confounders, then conditioning on the actions will not lead to self-delusions. In practice, there are many domains which could not be significantly affected by self-delusions. For instance, if the observations are sufficiently informative (with respect to the latent task and expert intentions), then the effect on the posterior over sequences can be reasonably assumed to be negligible.

\appendix
%%%%%%%%%%%%%%%%%%%%%%%%%%%%%%%%%%%%%%%%%%%%%%%%%%%%%%%%%%%%%%%
\section{Why does factual/counterfactual teaching work?}\label{sec:fcf-teaching}
%%%%%%%%%%%%%%%%%%%%%%%%%%%%%%%%%%%%%%%%%%%%%%%%%%%%%%%%%%%%%%%

We explain why the computation graph described in Figure~\ref{fig:computation-graph} leads to the regression of the correctly-conditioned and -intervened conditional probabilities. During training, each sample trajectory $\theta, a_1, \bar{a}_1, o_1, a_2, \bar{a}_2, o_2, \ldots, a_T, \bar{a}_T, o_T$ is generated by taking turns between the agent, the expert, and the task. For simplicity we assume that all the alphabets are finite and that the function approximator implementing the agent can represent the optimal solution. The first symbol~$\theta$ is unobserved by the agent but observed by the task and the expert. The observations $o_t$ and the expert actions $\bar{a}_t$ are used to compute losses, but the expert actions are immediately discarded thereafter and do not influence the remaining course of the realization. The probability of such a trajectory is
\[
  B(\theta, a_{\leq T}, \bar{a}_{\leq T}, o_{\leq T}) :=
  Q(\theta)
    \biggl( \prod_{\tau=1}^T 
      P(A_\tau = a_\tau \mid a_{<\tau}, o_{<\tau})
      Q(A_\tau = \bar{a}_\tau \mid \theta, a_{<\tau}, o_{<\tau})
      Q(O_\tau = o_\tau \mid \theta, a_{\leq \tau}, o_{<\tau})
    \biggr),
\]
where~$B$ stands for ``braided'' distribution, $P$ are the agent's probabilities and $Q$ are the environmental/expert probabilities. The probabilities of the right-hand-side correspond to the causal mechanisms. At time $t$, the agent gets two training signals, namely a factual and counterfactual one, given by
\[
  \ell(o_t) := - \log P(O_t = o_t \mid a_{\leq t}, o_{<t})
  \qquad\text{and}\qquad
  \ell(\bar{a}_t) := - \log P(A_t = \bar{a}_t \mid a_{<t}, o_{<t}),
\]
obtained with probabilities
\[
  \sum_\theta 
  B(\theta, a_{\leq t}, o_{<t})
  Q(O_t = o_t \mid \theta, a_{\leq t}, o_{<t})
  \qquad\text{and}\qquad
  \sum_\theta
  B(\theta, a_{<t}, o_{<t})
  Q(A_t = \bar{a}_t \mid \theta, a_{<t}, o_{<t})
\]
respectively. Note that the marginal distributions on $B$ were derived from the joint in the usual way, and that we average over the latent parameter $\Theta$ because it is unobserved by the agent.

The training setup implies that each loss term is optimized only w.r.t.\ a subset of the conditional probabilities. If we use backpropagation to optimize the loss in an end-to-end fashion, then the computation graph in Figure~\ref{fig:computation-graph}c (page~\pageref{fig:computation-graph}) implies that direct links such as $P_A \rightarrow \ell(\bar{A})$ propagate gradients, but stochastic links such as $P_A \rightsquigarrow A$ don't. By tracing the gradient paths originating in each loss, one sees that the minimization of the log-losses $\ell(\bar{a}_t)$ and $\ell(o_t)$ at time~$t$ \emph{only depend} on the conditional probabilities $P(A_t = \bar{a}_t \mid a_{<t}, o_{<t})$ for all $\bar{a}_t$ and $P(O_t = o_t \mid a_{\leq t}, o_{<t})$ for all $o_t$ respectively. Therefore, it is sufficient to analyze the log-losses separately and w.r.t.\ to given fixed histories $a_{<t}, o_{<t}$ averaged over $\theta$. Let us investigate actions and observations in turn.

For an action at time $t$, we have the expected log-loss:
\begin{align*}
  L(A_t) 
  &:= \sum_{\bar{a}_t} \sum_\theta
    B(\theta, a_{<t}, o_{<t})
    Q(\bar{a}_t \mid \theta, a_{<t}, o_{<t})
    \ell(\bar{a}_t) \\
  &\stackrel{(a)}{=} - \sum_{\bar{a}_t} \sum_\theta
    Q(\theta)
    \biggl( \prod_{\tau=1}^{t-1} 
      P(a_\tau \mid a_{<\tau}, o_{<\tau})
      Q(o_\tau \mid \theta, a_{\leq \tau}, o_{<\tau})
    \biggr)
    Q(\bar{a}_t \mid \theta, a_{<t}, o_{<t})
    \log P(\bar{a}_t \mid a_{<t}, o_{<t}) \\
  &\stackrel{(b)}{=} - \sum_{\bar{a}_t} \sum_\theta
    Q(\theta \mid \hat{a}_{<t}) 
    \biggl( \prod_{\tau=1}^{t-1} 
      P(a_\tau \mid a_{<\tau}, o_{<\tau})
      Q(o_\tau \mid \theta, \hat{a}_{<t}, o_{<\tau})
    \biggr)
    Q(\bar{a}_t \mid \theta, \hat{a}_{<t}, o_{<t})
    \log P(\bar{a}_t \mid a_{<t}, o_{<t}) \\
  &\stackrel{(c)}{=} 
    \biggl( \prod_{\tau=1}^{t-1} 
      P(a_\tau \mid a_{<\tau}, o_{<\tau})
      Q(o_\tau \mid \hat{a}_{<t}, o_{<\tau})
    \biggr)
    \biggl\{ -\sum_{\bar{a}_t}
    Q(\bar{a}_t \mid \hat{a}_{<t}, o_{<t})
    \log P(\bar{a}_t \mid a_{<t}, o_{<t}) \biggr\}.
\end{align*}
Equality~(a) is obtained by substituting the definitions.  Equality~(b) is obtained in two steps. First we convert the action conditions~$a_\tau$ to action interventions~$\hat{a}_\tau := \DO(a)_{\tau}$ in the $Q$ probabilities. This works because the $Q$ probabilities are the causal factors. Then we introduce the remaining action interventions up to time $t-1$ to get $\hat{a}_{<t} = \DO(a_{<t})$. One can do this because the probability arguments are independent of the actions downstream \emph{after} intervention. Finally, equality~(c) results from using the chain rule on the $Q$-terms to obtain $Q(\theta, \DO(a_{<t}), o_{<t})$, then marginalizing out $\theta$, and finally using the chain rule again to break the joint probabilities into conditional probabilities, plus a cosmetic re-grouping of the terms. 

Rewritten in this way reveals that $L(A_t)$ is an expectation over a collection of cross-entropies between $Q(A_t \mid \DO({a}_{<t}), o_{<t})$ and $P(A_t \mid a_{<t}, o_{<t})$, where the expectation is taken over all histories of length $t-1$. Hence, minimizing $L(A_t)$ w.r.t.\ $P(A_t \mid a_{<t}, o_{<t})$ implies
\begin{equation}\label{eq:fcf-action}
  P(a_t \mid a_{<t}, o_{<t}) 
  = Q(a_t \mid \DO(a_{<t}), o_{<t})
\end{equation}
for all histories $a_{<t}, o_{<t}$ of length $t-1$. In particular, note how $Q(a_t \mid \DO(a_{<t}), o_{<t})$ contains \emph{intervened} past actions in the conditional.

Following an analogous argument for the observation at time $t$, we get:
\begin{align*}
  L(O_t)
  &:=\sum_{o_t} \sum_\theta 
    B(\theta, a_{<t}, o_{<t})
    Q(A_t = \bar{a}_t \mid \theta, a_{<t}, o_{<t})
    \ell(o_t) \\
  &= - \sum_{o_t} \sum_\theta 
    Q(\theta)
    \biggl( \prod_{\tau=1}^t
      P(a_\tau \mid a_{<\tau}, o_{<\tau})
      Q(o_\tau \mid \theta, a_{\leq \tau}, o_{<\tau})
    \biggr)
    \log P(o_t \mid a_{\leq t}, o_{<t}) \\
  &= - \sum_{o_t} \sum_\theta 
    \biggl( \prod_{\tau=1}^t 
      P(a_\tau \mid a_{<\tau}, o_{<\tau})
      Q(o_\tau \mid \theta, \hat{a}_{\leq t}, o_{<\tau})
    \biggr)
    \log P(o_t \mid a_{\leq t}, o_{<t}) \\
  &= \biggl( \prod_{\tau=1}^{t-1} 
      P(a_\tau \mid a_{<\tau}, o_{<\tau})
      Q(o_\tau \mid \hat{a}_{\leq t}, o_{<\tau})
    \biggr)
    P(a_t \mid a_{<t}, o_{<t})
    \biggl\{ -\sum_{o_t}
    Q(o_t \mid \hat{a}_{\leq t}, o_{<t})
    \log P(o_t \mid a_{\leq t}, o_{<t}) \biggr\}.
\end{align*}
As before, minimizing $L(O_t)$ w.r.t.\ $P(o_t \mid a_{\leq t}, o_{<t})$ implies
\begin{equation}\label{eq:fcf-observation}
  P(o_t \mid a_{\leq t}, o_{<t})
  = Q(o_t \mid \DO(a_{\leq t}), o_{<t}).
\end{equation}
Hence, \eqref{eq:fcf-action} and \eqref{eq:fcf-observation} together show that the use of factual and counterfactual loss signals leads to conditional probabilities that correctly treat past actions as interventions.

\bibliography{main}

\end{document}

%% file: figures/boxes.pdf_tex
%% Creator: Inkscape 1.0.2 (e86c870879, 2021-01-15), www.inkscape.org
%% Accompanies image file 'boxes.pdf' (pdf, eps, ps)
\begingroup%
  \makeatletter%
  \providecommand\color[2][]{%
    \errmessage{(Inkscape) Color is used for the text in Inkscape, but the package 'color.sty' is not loaded}%
    \renewcommand\color[2][]{}%
  }%
  \providecommand\transparent[1]{%
    \errmessage{(Inkscape) Transparency is used (non-zero) for the text in Inkscape, but the package 'transparent.sty' is not loaded}%
    \renewcommand\transparent[1]{}%
  }%
  \providecommand\rotatebox[2]{#2}%
  \newcommand*\fsize{\dimexpr\f@size pt\relax}%
  \newcommand*\lineheight[1]{\fontsize{\fsize}{#1\fsize}\selectfont}%
  \ifx\svgwidth\undefined%
    \setlength{\unitlength}{360.75100636bp}%
    \ifx\svgscale\undefined%
      \relax%
    \else%
      \setlength{\unitlength}{\unitlength * \real{\svgscale}}%
    \fi%
  \else%
    \setlength{\unitlength}{\svgwidth}%
  \fi%
  \global\let\svgwidth\undefined%
  \global\let\svgscale\undefined%
  \makeatother%
  \begin{picture}(1,0.35567482)%
    \lineheight{1}%
    \setlength\tabcolsep{0pt}%
    \put(0,0){\includegraphics[width=\unitlength,page=1]{boxes.pdf}}%
    \put(0.16501008,0.32424433){\makebox(0,0)[t]{\lineheight{1.25}\smash{\begin{tabular}[t]{c}$\Theta = 1$\end{tabular}}}}%
    \put(0.55789127,0.32424433){\makebox(0,0)[t]{\lineheight{1.25}\smash{\begin{tabular}[t]{c}$\Theta = 2$\end{tabular}}}}%
    \put(0,0){\includegraphics[width=\unitlength,page=2]{boxes.pdf}}%
    \put(0.55789127,0.00993938){\makebox(0,0)[t]{\lineheight{1.25}\smash{\begin{tabular}[t]{c}$A$\end{tabular}}}}%
    \put(0.667898,0.00993938){\makebox(0,0)[t]{\lineheight{1.25}\smash{\begin{tabular}[t]{c}$O$\end{tabular}}}}%
    \put(0.16501008,0.00993938){\makebox(0,0)[t]{\lineheight{1.25}\smash{\begin{tabular}[t]{c}$A$\end{tabular}}}}%
    \put(0.27501681,0.00993938){\makebox(0,0)[t]{\lineheight{1.25}\smash{\begin{tabular}[t]{c}$O$\end{tabular}}}}%
    \put(0.00785761,0.32424433){\makebox(0,0)[lt]{\lineheight{1.25}\smash{\begin{tabular}[t]{l}a)\end{tabular}}}}%
    \put(0.79361998,0.32424433){\makebox(0,0)[lt]{\lineheight{1.25}\smash{\begin{tabular}[t]{l}b)\end{tabular}}}}%
    \put(0.33409155,0.26901726){\makebox(0,0)[rt]{\lineheight{1.25}\smash{\begin{tabular}[t]{r}+1\end{tabular}}}}%
    \put(0.33409155,0.20615627){\makebox(0,0)[rt]{\lineheight{1.25}\smash{\begin{tabular}[t]{r}-1\end{tabular}}}}%
    \put(0.33304838,0.12758004){\makebox(0,0)[rt]{\lineheight{1.25}\smash{\begin{tabular}[t]{r}+1\end{tabular}}}}%
    \put(0.33409155,0.06471905){\makebox(0,0)[rt]{\lineheight{1.25}\smash{\begin{tabular}[t]{r}-1\end{tabular}}}}%
    \put(0.19265432,0.19066473){\makebox(0,0)[rt]{\lineheight{1.25}\smash{\begin{tabular}[t]{r}1\end{tabular}}}}%
    \put(0.19265432,0.0492275){\makebox(0,0)[rt]{\lineheight{1.25}\smash{\begin{tabular}[t]{r}2\end{tabular}}}}%
    \put(0.72697274,0.26879356){\makebox(0,0)[rt]{\lineheight{1.25}\smash{\begin{tabular}[t]{r}+1\end{tabular}}}}%
    \put(0.72697274,0.20593257){\makebox(0,0)[rt]{\lineheight{1.25}\smash{\begin{tabular}[t]{r}-1\end{tabular}}}}%
    \put(0.72592957,0.12735634){\makebox(0,0)[rt]{\lineheight{1.25}\smash{\begin{tabular}[t]{r}+1\end{tabular}}}}%
    \put(0.72697274,0.06449535){\makebox(0,0)[rt]{\lineheight{1.25}\smash{\begin{tabular}[t]{r}-1\end{tabular}}}}%
    \put(0.58553551,0.19044103){\makebox(0,0)[rt]{\lineheight{1.25}\smash{\begin{tabular}[t]{r}1\end{tabular}}}}%
    \put(0.58553551,0.0490038){\makebox(0,0)[rt]{\lineheight{1.25}\smash{\begin{tabular}[t]{r}2\end{tabular}}}}%
    \put(0.9664877,0.13359777){\makebox(0,0)[t]{\lineheight{1.25}\smash{\begin{tabular}[t]{c}$O$\end{tabular}}}}%
    \put(0.85648097,0.13359777){\makebox(0,0)[t]{\lineheight{1.25}\smash{\begin{tabular}[t]{c}$A$\end{tabular}}}}%
    \put(0.91148434,0.22788925){\makebox(0,0)[t]{\lineheight{1.25}\smash{\begin{tabular}[t]{c}$\Theta$\end{tabular}}}}%
  \end{picture}%
\endgroup%

%% file: figures/boxes-dag.pdf_tex
%% Creator: Inkscape 1.0.2 (e86c870879, 2021-01-15), www.inkscape.org
%% Accompanies image file 'boxes-dag.pdf' (pdf, eps, ps)
\begingroup%
  \makeatletter%
  \providecommand\color[2][]{%
    \errmessage{(Inkscape) Color is used for the text in Inkscape, but the package 'color.sty' is not loaded}%
    \renewcommand\color[2][]{}%
  }%
  \providecommand\transparent[1]{%
    \errmessage{(Inkscape) Transparency is used (non-zero) for the text in Inkscape, but the package 'transparent.sty' is not loaded}%
    \renewcommand\transparent[1]{}%
  }%
  \providecommand\rotatebox[2]{#2}%
  \newcommand*\fsize{\dimexpr\f@size pt\relax}%
  \newcommand*\lineheight[1]{\fontsize{\fsize}{#1\fsize}\selectfont}%
  \ifx\svgwidth\undefined%
    \setlength{\unitlength}{323.90061266bp}%
    \ifx\svgscale\undefined%
      \relax%
    \else%
      \setlength{\unitlength}{\unitlength * \real{\svgscale}}%
    \fi%
  \else%
    \setlength{\unitlength}{\svgwidth}%
  \fi%
  \global\let\svgwidth\undefined%
  \global\let\svgscale\undefined%
  \makeatother%
  \begin{picture}(1,0.23861155)%
    \lineheight{1}%
    \setlength\tabcolsep{0pt}%
    \put(0,0){\includegraphics[width=\unitlength,page=1]{boxes-dag.pdf}}%
    \put(0.16761286,0.07819803){\makebox(0,0)[t]{\lineheight{1.25}\smash{\begin{tabular}[t]{c}$O$\end{tabular}}}}%
    \put(0.04509059,0.07819803){\makebox(0,0)[t]{\lineheight{1.25}\smash{\begin{tabular}[t]{c}$A$\end{tabular}}}}%
    \put(0.10635173,0.18321712){\makebox(0,0)[t]{\lineheight{1.25}\smash{\begin{tabular}[t]{c}$\Theta$\end{tabular}}}}%
    \put(0,0){\includegraphics[width=\unitlength,page=2]{boxes-dag.pdf}}%
    \put(0.10617839,0.0186625){\makebox(0,0)[t]{\lineheight{1.25}\smash{\begin{tabular}[t]{c}$P(O \mid A)$\end{tabular}}}}%
    \put(0.36872612,0.0186625){\makebox(0,0)[t]{\lineheight{1.25}\smash{\begin{tabular}[t]{c}$P(O \mid \text{do}(A))$\end{tabular}}}}%
    \put(0,0){\includegraphics[width=\unitlength,page=3]{boxes-dag.pdf}}%
    \put(0.42998726,0.07762521){\makebox(0,0)[t]{\lineheight{1.25}\smash{\begin{tabular}[t]{c}$O$\end{tabular}}}}%
    \put(0.30746498,0.07762521){\makebox(0,0)[t]{\lineheight{1.25}\smash{\begin{tabular}[t]{c}$A$\end{tabular}}}}%
    \put(0.36872612,0.18264431){\makebox(0,0)[t]{\lineheight{1.25}\smash{\begin{tabular}[t]{c}$\Theta$\end{tabular}}}}%
    \put(0,0){\includegraphics[width=\unitlength,page=4]{boxes-dag.pdf}}%
    \put(0.69253499,0.07762528){\makebox(0,0)[t]{\lineheight{1.25}\smash{\begin{tabular}[t]{c}$O$\end{tabular}}}}%
    \put(0.57001271,0.07762528){\makebox(0,0)[t]{\lineheight{1.25}\smash{\begin{tabular}[t]{c}$A$\end{tabular}}}}%
    \put(0.63127385,0.18264437){\makebox(0,0)[t]{\lineheight{1.25}\smash{\begin{tabular}[t]{c}$\Theta$\end{tabular}}}}%
    \put(0.63127385,0.01866248){\color[rgb]{0,0,0}\makebox(0,0)[t]{\lineheight{1.25}\smash{\begin{tabular}[t]{c}$P(O \mid \Theta, A)$\end{tabular}}}}%
    \put(0.00991089,0.18494271){\makebox(0,0)[lt]{\lineheight{1.25}\smash{\begin{tabular}[t]{l}a)\end{tabular}}}}%
    \put(0.27245862,0.18494271){\makebox(0,0)[lt]{\lineheight{1.25}\smash{\begin{tabular}[t]{l}b)\end{tabular}}}}%
    \put(0.53500635,0.18494271){\makebox(0,0)[lt]{\lineheight{1.25}\smash{\begin{tabular}[t]{l}c)\end{tabular}}}}%
    \put(0,0){\includegraphics[width=\unitlength,page=5]{boxes-dag.pdf}}%
    \put(0.95508251,0.07762528){\makebox(0,0)[t]{\lineheight{1.25}\smash{\begin{tabular}[t]{c}$O$\end{tabular}}}}%
    \put(0.83256031,0.07762528){\makebox(0,0)[t]{\lineheight{1.25}\smash{\begin{tabular}[t]{c}$A$\end{tabular}}}}%
    \put(0.89382138,0.18264437){\makebox(0,0)[t]{\lineheight{1.25}\smash{\begin{tabular}[t]{c}$\Theta$\end{tabular}}}}%
    \put(0.89382138,0.01866248){\color[rgb]{0,0,0}\makebox(0,0)[t]{\lineheight{1.25}\smash{\begin{tabular}[t]{c}$P(O \mid \Theta, \text{do}(A))$\end{tabular}}}}%
    \put(0.76254771,0.12368157){\color[rgb]{0,0,0}\makebox(0,0)[t]{\lineheight{1.25}\smash{\begin{tabular}[t]{c}=\end{tabular}}}}%
  \end{picture}%
\endgroup%

%% file: figures/sequential-game.pdf_tex
%% Creator: Inkscape 1.0.2 (e86c870879, 2021-01-15), www.inkscape.org
%% Accompanies image file 'sequential-game.pdf' (pdf, eps, ps)
\begingroup%
  \makeatletter%
  \providecommand\color[2][]{%
    \errmessage{(Inkscape) Color is used for the text in Inkscape, but the package 'color.sty' is not loaded}%
    \renewcommand\color[2][]{}%
  }%
  \providecommand\transparent[1]{%
    \errmessage{(Inkscape) Transparency is used (non-zero) for the text in Inkscape, but the package 'transparent.sty' is not loaded}%
    \renewcommand\transparent[1]{}%
  }%
  \providecommand\rotatebox[2]{#2}%
  \newcommand*\fsize{\dimexpr\f@size pt\relax}%
  \newcommand*\lineheight[1]{\fontsize{\fsize}{#1\fsize}\selectfont}%
  \ifx\svgwidth\undefined%
    \setlength{\unitlength}{479.80608122bp}%
    \ifx\svgscale\undefined%
      \relax%
    \else%
      \setlength{\unitlength}{\unitlength * \real{\svgscale}}%
    \fi%
  \else%
    \setlength{\unitlength}{\svgwidth}%
  \fi%
  \global\let\svgwidth\undefined%
  \global\let\svgscale\undefined%
  \makeatother%
  \begin{picture}(1,0.40921023)%
    \lineheight{1}%
    \setlength\tabcolsep{0pt}%
    \put(0,0){\includegraphics[width=\unitlength,page=1]{sequential-game.pdf}}%
    \put(0.32163237,0.01968608){\makebox(0,0)[t]{\lineheight{1.25}\smash{\begin{tabular}[t]{c}$O_2$\end{tabular}}}}%
    \put(0,0){\includegraphics[width=\unitlength,page=2]{sequential-game.pdf}}%
    \put(0.26255338,0.01968608){\makebox(0,0)[t]{\lineheight{1.25}\smash{\begin{tabular}[t]{c}$A_2$\end{tabular}}}}%
    \put(0,0){\includegraphics[width=\unitlength,page=3]{sequential-game.pdf}}%
    \put(0.20347439,0.01968608){\makebox(0,0)[t]{\lineheight{1.25}\smash{\begin{tabular}[t]{c}$\Theta_2$\end{tabular}}}}%
    \put(0,0){\includegraphics[width=\unitlength,page=4]{sequential-game.pdf}}%
    \put(0.14439541,0.01968608){\makebox(0,0)[t]{\lineheight{1.25}\smash{\begin{tabular}[t]{c}$O_1$\end{tabular}}}}%
    \put(0,0){\includegraphics[width=\unitlength,page=5]{sequential-game.pdf}}%
    \put(0.08531642,0.01968608){\makebox(0,0)[t]{\lineheight{1.25}\smash{\begin{tabular}[t]{c}$A_1$\end{tabular}}}}%
    \put(0,0){\includegraphics[width=\unitlength,page=6]{sequential-game.pdf}}%
    \put(0.02623744,0.01968608){\makebox(0,0)[t]{\lineheight{1.25}\smash{\begin{tabular}[t]{c}$\Theta_1$\end{tabular}}}}%
    \put(0,0){\includegraphics[width=\unitlength,page=7]{sequential-game.pdf}}%
    \put(0.69382976,0.01968617){\makebox(0,0)[t]{\lineheight{1.25}\smash{\begin{tabular}[t]{c}$O_2$\end{tabular}}}}%
    \put(0,0){\includegraphics[width=\unitlength,page=8]{sequential-game.pdf}}%
    \put(0.63475078,0.01968617){\makebox(0,0)[t]{\lineheight{1.25}\smash{\begin{tabular}[t]{c}$A_2$\end{tabular}}}}%
    \put(0,0){\includegraphics[width=\unitlength,page=9]{sequential-game.pdf}}%
    \put(0.57567179,0.01968626){\makebox(0,0)[t]{\lineheight{1.25}\smash{\begin{tabular}[t]{c}$\Theta_2$\end{tabular}}}}%
    \put(0,0){\includegraphics[width=\unitlength,page=10]{sequential-game.pdf}}%
    \put(0.5165928,0.01968617){\makebox(0,0)[t]{\lineheight{1.25}\smash{\begin{tabular}[t]{c}$O_1$\end{tabular}}}}%
    \put(0,0){\includegraphics[width=\unitlength,page=11]{sequential-game.pdf}}%
    \put(0.45751382,0.01968617){\makebox(0,0)[t]{\lineheight{1.25}\smash{\begin{tabular}[t]{c}$A_1$\end{tabular}}}}%
    \put(0,0){\includegraphics[width=\unitlength,page=12]{sequential-game.pdf}}%
    \put(0.39843483,0.01968617){\makebox(0,0)[t]{\lineheight{1.25}\smash{\begin{tabular}[t]{c}$\Theta_1$\end{tabular}}}}%
    \put(0,0){\includegraphics[width=\unitlength,page=13]{sequential-game.pdf}}%
    \put(0.11894059,0.25482227){\makebox(0,0)[t]{\lineheight{1.25}\smash{\begin{tabular}[t]{c}$\Theta_2$\end{tabular}}}}%
    \put(0.11894059,0.19574328){\makebox(0,0)[t]{\lineheight{1.25}\smash{\begin{tabular}[t]{c}$O_2$\end{tabular}}}}%
    \put(0.11894059,0.22528277){\makebox(0,0)[t]{\lineheight{1.25}\smash{\begin{tabular}[t]{c}$A_2$\end{tabular}}}}%
    \put(0.11894059,0.28436176){\makebox(0,0)[t]{\lineheight{1.25}\smash{\begin{tabular}[t]{c}$O_1$\end{tabular}}}}%
    \put(0.11894059,0.31390125){\makebox(0,0)[t]{\lineheight{1.25}\smash{\begin{tabular}[t]{c}$A_1$\end{tabular}}}}%
    \put(0.11894059,0.34344075){\makebox(0,0)[t]{\lineheight{1.25}\smash{\begin{tabular}[t]{c}$\Theta_1$\end{tabular}}}}%
    \put(0.1012169,0.37888811){\makebox(0,0)[t]{\lineheight{1.25}\smash{\begin{tabular}[t]{c}a)\end{tabular}}}}%
    \put(0.03623001,0.14257219){\makebox(0,0)[t]{\lineheight{1.25}\smash{\begin{tabular}[t]{c}b)\end{tabular}}}}%
    \put(0.40842762,0.14257219){\makebox(0,0)[t]{\lineheight{1.25}\smash{\begin{tabular}[t]{c}c)\end{tabular}}}}%
    \put(0,0){\includegraphics[width=\unitlength,page=14]{sequential-game.pdf}}%
    \put(0.76880944,0.01850628){\makebox(0,0)[t]{\lineheight{1.25}\smash{\begin{tabular}[t]{c}$O_2$\end{tabular}}}}%
    \put(0.76880944,0.04804577){\makebox(0,0)[t]{\lineheight{1.25}\smash{\begin{tabular}[t]{c}$A_2$\end{tabular}}}}%
    \put(0.76880944,0.07758531){\makebox(0,0)[t]{\lineheight{1.25}\smash{\begin{tabular}[t]{c}$O_1$\end{tabular}}}}%
    \put(0.76880944,0.1071248){\makebox(0,0)[t]{\lineheight{1.25}\smash{\begin{tabular}[t]{c}$A_1$\end{tabular}}}}%
    \put(0.75108574,0.14848009){\makebox(0,0)[t]{\lineheight{1.25}\smash{\begin{tabular}[t]{c}d)\end{tabular}}}}%
    \put(0,0){\includegraphics[width=\unitlength,page=15]{sequential-game.pdf}}%
  \end{picture}%
\endgroup%

%% file: figures/computation-graph.pdf_tex
%% Creator: Inkscape 1.0.2 (e86c870879, 2021-01-15), www.inkscape.org
%% Accompanies image file 'computation-graph.pdf' (pdf, eps, ps)
\begingroup%
  \makeatletter%
  \providecommand\color[2][]{%
    \errmessage{(Inkscape) Color is used for the text in Inkscape, but the package 'color.sty' is not loaded}%
    \renewcommand\color[2][]{}%
  }%
  \providecommand\transparent[1]{%
    \errmessage{(Inkscape) Transparency is used (non-zero) for the text in Inkscape, but the package 'transparent.sty' is not loaded}%
    \renewcommand\transparent[1]{}%
  }%
  \providecommand\rotatebox[2]{#2}%
  \newcommand*\fsize{\dimexpr\f@size pt\relax}%
  \newcommand*\lineheight[1]{\fontsize{\fsize}{#1\fsize}\selectfont}%
  \ifx\svgwidth\undefined%
    \setlength{\unitlength}{454.29422694bp}%
    \ifx\svgscale\undefined%
      \relax%
    \else%
      \setlength{\unitlength}{\unitlength * \real{\svgscale}}%
    \fi%
  \else%
    \setlength{\unitlength}{\svgwidth}%
  \fi%
  \global\let\svgwidth\undefined%
  \global\let\svgscale\undefined%
  \makeatother%
  \begin{picture}(1,0.28243824)%
    \lineheight{1}%
    \setlength\tabcolsep{0pt}%
    \put(0,0){\includegraphics[width=\unitlength,page=1]{computation-graph.pdf}}%
    \put(0.50012406,0.24294337){\makebox(0,0)[t]{\lineheight{1.25}\smash{\begin{tabular}[t]{c}$E'$\end{tabular}}}}%
    \put(0,0){\includegraphics[width=\unitlength,page=2]{computation-graph.pdf}}%
    \put(0.50012406,0.14934836){\makebox(0,0)[t]{\lineheight{1.25}\smash{\begin{tabular}[t]{c}$W'$\end{tabular}}}}%
    \put(0,0){\includegraphics[width=\unitlength,page=3]{computation-graph.pdf}}%
    \put(0.50012406,0.02455498){\makebox(0,0)[t]{\lineheight{1.25}\smash{\begin{tabular}[t]{c}$M'$\end{tabular}}}}%
    \put(0,0){\includegraphics[width=\unitlength,page=4]{computation-graph.pdf}}%
    \put(0.40652902,0.08695167){\color[rgb]{1,0,0}\makebox(0,0)[t]{\lineheight{1.25}\smash{\begin{tabular}[t]{c}$\ell$\end{tabular}}}}%
    \put(0,0){\includegraphics[width=\unitlength,page=5]{computation-graph.pdf}}%
    \put(0.34413233,0.11815001){\makebox(0,0)[t]{\lineheight{1.25}\smash{\begin{tabular}[t]{c}$O$\end{tabular}}}}%
    \put(0,0){\includegraphics[width=\unitlength,page=6]{computation-graph.pdf}}%
    \put(0.34413252,0.05575337){\makebox(0,0)[t]{\lineheight{1.25}\smash{\begin{tabular}[t]{c}$P_O$\end{tabular}}}}%
    \put(0,0){\includegraphics[width=\unitlength,page=7]{computation-graph.pdf}}%
    \put(0.28173564,0.24294337){\makebox(0,0)[t]{\lineheight{1.25}\smash{\begin{tabular}[t]{c}$E$\end{tabular}}}}%
    \put(0,0){\includegraphics[width=\unitlength,page=8]{computation-graph.pdf}}%
    \put(0.28173564,0.14934836){\makebox(0,0)[t]{\lineheight{1.25}\smash{\begin{tabular}[t]{c}$W$\end{tabular}}}}%
    \put(0,0){\includegraphics[width=\unitlength,page=9]{computation-graph.pdf}}%
    \put(0.28173564,0.02455498){\makebox(0,0)[t]{\lineheight{1.25}\smash{\begin{tabular}[t]{c}$M$\end{tabular}}}}%
    \put(0,0){\includegraphics[width=\unitlength,page=10]{computation-graph.pdf}}%
    \put(0.93690071,0.2429434){\makebox(0,0)[t]{\lineheight{1.25}\smash{\begin{tabular}[t]{c}$E'$\end{tabular}}}}%
    \put(0,0){\includegraphics[width=\unitlength,page=11]{computation-graph.pdf}}%
    \put(0.93690071,0.14934836){\makebox(0,0)[t]{\lineheight{1.25}\smash{\begin{tabular}[t]{c}$W'$\end{tabular}}}}%
    \put(0,0){\includegraphics[width=\unitlength,page=12]{computation-graph.pdf}}%
    \put(0.93690071,0.02455498){\makebox(0,0)[t]{\lineheight{1.25}\smash{\begin{tabular}[t]{c}$M'$\end{tabular}}}}%
    \put(0,0){\includegraphics[width=\unitlength,page=13]{computation-graph.pdf}}%
    \put(0.78090898,0.08695167){\color[rgb]{1,0,0}\makebox(0,0)[t]{\lineheight{1.25}\smash{\begin{tabular}[t]{c}$\ell$\end{tabular}}}}%
    \put(0,0){\includegraphics[width=\unitlength,page=14]{computation-graph.pdf}}%
    \put(0.71908507,0.21077931){\makebox(0,0)[t]{\lineheight{1.25}\smash{\begin{tabular}[t]{c}$\bar{A}$\end{tabular}}}}%
    \put(0,0){\includegraphics[width=\unitlength,page=15]{computation-graph.pdf}}%
    \put(0.6561156,0.2429434){\makebox(0,0)[t]{\lineheight{1.25}\smash{\begin{tabular}[t]{c}$E$\end{tabular}}}}%
    \put(0,0){\includegraphics[width=\unitlength,page=16]{computation-graph.pdf}}%
    \put(0.6561156,0.14934836){\makebox(0,0)[t]{\lineheight{1.25}\smash{\begin{tabular}[t]{c}$W$\end{tabular}}}}%
    \put(0,0){\includegraphics[width=\unitlength,page=17]{computation-graph.pdf}}%
    \put(0.6561156,0.02455498){\makebox(0,0)[t]{\lineheight{1.25}\smash{\begin{tabular}[t]{c}$M$\end{tabular}}}}%
    \put(0,0){\includegraphics[width=\unitlength,page=18]{computation-graph.pdf}}%
    \put(0.12574372,0.24294344){\makebox(0,0)[t]{\lineheight{1.25}\smash{\begin{tabular}[t]{c}$E$\end{tabular}}}}%
    \put(0,0){\includegraphics[width=\unitlength,page=19]{computation-graph.pdf}}%
    \put(0.12574372,0.14934841){\makebox(0,0)[t]{\lineheight{1.25}\smash{\begin{tabular}[t]{c}$W$\end{tabular}}}}%
    \put(0,0){\includegraphics[width=\unitlength,page=20]{computation-graph.pdf}}%
    \put(0.12574372,0.02455502){\makebox(0,0)[t]{\lineheight{1.25}\smash{\begin{tabular}[t]{c}$M$\end{tabular}}}}%
    \put(0,0){\includegraphics[width=\unitlength,page=21]{computation-graph.pdf}}%
    \put(0.03214868,0.19926574){\makebox(0,0)[t]{\lineheight{1.25}\smash{\begin{tabular}[t]{c}$\Theta$\end{tabular}}}}%
    \put(0,0){\includegraphics[width=\unitlength,page=22]{computation-graph.pdf}}%
    \put(0.01330609,0.25041337){\makebox(0,0)[lt]{\lineheight{1.25}\smash{\begin{tabular}[t]{l}a)\end{tabular}}}}%
    \put(0.20673583,0.25041337){\makebox(0,0)[lt]{\lineheight{1.25}\smash{\begin{tabular}[t]{l}b)\end{tabular}}}}%
    \put(0.58111579,0.25041337){\makebox(0,0)[lt]{\lineheight{1.25}\smash{\begin{tabular}[t]{l}c)\end{tabular}}}}%
    \put(0.02578543,0.05074395){\rotatebox{90}{\makebox(0,0)[t]{\lineheight{1.25}\smash{\begin{tabular}[t]{c}Agent\end{tabular}}}}}%
    \put(0,0){\includegraphics[width=\unitlength,page=23]{computation-graph.pdf}}%
    \put(0.30657073,0.05698362){\makebox(0,0)[t]{\lineheight{1.25}\smash{\begin{tabular}[t]{c}$f_O$\end{tabular}}}}%
    \put(0.39392591,0.01330594){\makebox(0,0)[t]{\lineheight{1.25}\smash{\begin{tabular}[t]{c}$g_O$\end{tabular}}}}%
    \put(0.6809505,0.05698362){\makebox(0,0)[t]{\lineheight{1.25}\smash{\begin{tabular}[t]{c}$f_A$\end{tabular}}}}%
    \put(0.79950421,0.01330594){\makebox(0,0)[t]{\lineheight{1.25}\smash{\begin{tabular}[t]{c}$g_A$\end{tabular}}}}%
    \put(0,0){\includegraphics[width=\unitlength,page=24]{computation-graph.pdf}}%
    \put(0.84330605,0.05575342){\makebox(0,0)[t]{\lineheight{1.25}\smash{\begin{tabular}[t]{c}$A$\end{tabular}}}}%
    \put(0.71851267,0.05575337){\makebox(0,0)[t]{\lineheight{1.25}\smash{\begin{tabular}[t]{c}$P_A$\end{tabular}}}}%
  \end{picture}%
\endgroup%

%% file: main.bbl
\begin{thebibliography}{10}

\bibitem{abbeel2004apprenticeship}
P.~Abbeel and A.~Y. Ng.
\newblock Apprenticeship learning via inverse reinforcement learning.
\newblock In {\em Proceedings of the twenty-first international conference on
  Machine learning}, page~1, 2004.

\bibitem{arjovsky2019invariant}
M.~Arjovsky, L.~Bottou, I.~Gulrajani, and D.~Lopez-Paz.
\newblock Invariant risk minimization.
\newblock {\em arXiv preprint arXiv:1907.02893}, 2019.

\bibitem{blakemore2000can}
S.-J. Blakemore, D.~Wolpert, and C.~Frith.
\newblock Why can't you tickle yourself?
\newblock {\em Neuroreport}, 11(11):R11--R16, 2000.

\bibitem{Bommasani:21foundation}
R.~Bommasani, D.~A. Hudson, E.~Adeli, R.~Altman, S.~Arora, S.~von Arx, M.~S.
  Bernstein, J.~Bohg, A.~Bosselut, E.~Brunskill, E.~Brynjolfsson, S.~Buch,
  D.~Card, R.~Castellon, N.~Chatterji, A.~S. Chen, K.~Creel, J.~Q. Davis,
  D.~Demszky, C.~Donahue, M.~Doumbouya, E.~Durmus, S.~Ermon, J.~Etchemendy,
  K.~Ethayarajh, L.~Fei{-}Fei, C.~Finn, T.~Gale, L.~Gillespie, K.~Goel, N.~D.
  Goodman, S.~Grossman, N.~Guha, T.~Hashimoto, P.~Henderson, J.~Hewitt, D.~E.
  Ho, J.~Hong, K.~Hsu, J.~Huang, T.~Icard, S.~Jain, D.~Jurafsky, P.~Kalluri,
  S.~Karamcheti, G.~Keeling, F.~Khani, O.~Khattab, P.~W. Koh, M.~S. Krass,
  R.~Krishna, R.~Kuditipudi, and et~al.
\newblock On the opportunities and risks of foundation models.
\newblock {\em arXiv preprint arXiv:2108.07258}, 2021.

\bibitem{brown2020language}
T.~B. Brown, B.~Mann, N.~Ryder, M.~Subbiah, J.~Kaplan, P.~Dhariwal,
  A.~Neelakantan, P.~Shyam, G.~Sastry, A.~Askell, et~al.
\newblock Language models are few-shot learners.
\newblock {\em arXiv preprint arXiv:2005.14165}, 2020.

\bibitem{chen2021decision}
L.~Chen, K.~Lu, A.~Rajeswaran, K.~Lee, A.~Grover, M.~Laskin, P.~Abbeel,
  A.~Srinivas, and I.~Mordatch.
\newblock Decision transformer: Reinforcement learning via sequence modeling.
\newblock {\em arXiv preprint arXiv:2106.01345}, 2021.

\bibitem{cover1999elements}
T.~M. Cover.
\newblock {\em Elements of information theory}.
\newblock John Wiley \& Sons, 1999.

\bibitem{deletang2021causal}
G.~D{\'e}letang, J.~Grau-Moya, M.~Martic, T.~Genewein, T.~McGrath, V.~Mikulik,
  M.~Kunesch, S.~Legg, and P.~A. Ortega.
\newblock Causal analysis of agent behavior for ai safety.
\newblock {\em arXiv preprint arXiv:2103.03938}, 2021.

\bibitem{duan2016rl}
Y.~Duan, J.~Schulman, X.~Chen, P.~L. Bartlett, I.~Sutskever, and P.~Abbeel.
\newblock Rl$^2$: Fast reinforcement learning via slow reinforcement learning.
\newblock {\em arXiv preprint arXiv:1611.02779}, 2016.

\bibitem{ho2016generative}
J.~Ho and S.~Ermon.
\newblock Generative adversarial imitation learning.
\newblock {\em Advances in neural information processing systems},
  29:4565--4573, 2016.

\bibitem{janner2021reinforcement}
M.~Janner, Q.~Li, and S.~Levine.
\newblock Reinforcement learning as one big sequence modeling problem.
\newblock {\em arXiv preprint arXiv:2106.02039}, 2021.

\bibitem{leike2016thompson}
J.~Leike, T.~Lattimore, L.~Orseau, and M.~Hutter.
\newblock Thompson sampling is asymptotically optimal in general environments.
\newblock In {\em Proceedings of the Thirty-Second Conference on Uncertainty in
  Artificial Intelligence, {UAI} 2016}. {AUAI} Press, 2016.

\bibitem{ng2000algorithms}
A.~Y. Ng, S.~J. Russell, et~al.
\newblock Algorithms for inverse reinforcement learning.
\newblock In {\em Icml}, volume~1, page~2, 2000.

\bibitem{ortega2010bayesian}
P.~Ortega and D.~Braun.
\newblock A {B}ayesian rule for adaptive control based on causal interventions.
\newblock In {\em Third Conference on Artificial General Intelligence (AGI
  2010)}, pages 121--126. Atlantis Press, 2010.

\bibitem{ortega2010minimum}
P.~A. Ortega and D.~A. Braun.
\newblock A minimum relative entropy principle for learning and acting.
\newblock {\em Journal of Artificial Intelligence Research}, 38:475--511, 2010.

\bibitem{osband2017posterior}
I.~Osband and B.~Van~Roy.
\newblock Why is posterior sampling better than optimism for reinforcement
  learning?
\newblock In {\em International conference on machine learning}, pages
  2701--2710. PMLR, 2017.

\bibitem{osborne1994course}
M.~J. Osborne and A.~Rubinstein.
\newblock {\em A course in game theory}.
\newblock MIT press, 1994.

\bibitem{pearl2009causality}
J.~Pearl.
\newblock {\em Causality}.
\newblock Cambridge university press, 2009.

\bibitem{rezende2020causally}
D.~J. Rezende, I.~Danihelka, G.~Papamakarios, N.~R. Ke, R.~Jiang, T.~Weber,
  K.~Gregor, H.~Merzic, F.~Viola, J.~Wang, et~al.
\newblock Causally correct partial models for reinforcement learning.
\newblock {\em arXiv preprint arXiv:2002.02836}, 2020.

\bibitem{ross2011reduction}
S.~Ross, G.~Gordon, and D.~Bagnell.
\newblock A reduction of imitation learning and structured prediction to
  no-regret online learning.
\newblock In {\em Proceedings of the fourteenth international conference on
  artificial intelligence and statistics}, pages 627--635. JMLR Workshop and
  Conference Proceedings, 2011.

\bibitem{sutton2018reinforcement}
R.~S. Sutton and A.~G. Barto.
\newblock {\em Reinforcement learning: An introduction}.
\newblock MIT press, 2018.

\bibitem{veness2015compress}
J.~Veness, M.~G. Bellemare, M.~Hutter, A.~Chua, and G.~Desjardins.
\newblock Compress and control.
\newblock In {\em Twenty-Ninth AAAI Conference on Artificial Intelligence},
  2015.

\bibitem{vonNeumann1947theory}
J.~Von~Neumann and O.~Morgenstern.
\newblock {\em Theory of games and economic behavior, 2nd rev}.
\newblock Princeton university press, 1947.

\bibitem{wang2016learning}
J.~X. Wang, Z.~Kurth-Nelson, D.~Tirumala, H.~Soyer, J.~Z. Leibo, R.~Munos,
  C.~Blundell, D.~Kumaran, and M.~Botvinick.
\newblock Learning to reinforcement learn.
\newblock {\em arXiv preprint arXiv:1611.05763}, 2016.

\bibitem{zolna2019task}
K.~Zolna, S.~Reed, A.~Novikov, S.~G. Colmenarejo, D.~Budden, S.~Cabi, M.~Denil,
  N.~de~Freitas, and Z.~Wang.
\newblock Task-relevant adversarial imitation learning.
\newblock {\em arXiv preprint arXiv:1910.01077}, 2019.

\end{thebibliography}
